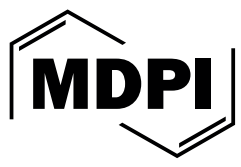

Article

# A Wearable Stiffness-Rendering Haptic Device with a Honeycomb Jamming Mechanism for Bilateral Teleoperation

**Thomas M. Kwok** , **Bohan Zhang** and **Wai Tuck Chow** *

School of Mechanical and Aerospace Engineering, Nanyang Technological University, Singapore 639798, Singapore; thomasm.kwok@u.nus.edu (T.M.K.); bzhang025@e.ntu.edu.sg (B.Z.)
* Correspondence: wtchow@ntu.edu.sg

**Abstract:** This paper addresses the challenge of providing kinesthetic feedback in bilateral teleoperation by designing a wearable, lightweight (20 g), and compact haptic device, the HJ-Haptic, utilizing a honeycomb jamming mechanism for object stiffness rendering. The HJ-Haptic device can vary its stiffness, from 1.15 N/mm to 2.64 N/mm, using a 30 kPa vacuum pressure. We demonstrate its implementation in a teleoperation framework, enabling operators to adjust grip force based on a reliable haptic feedback on object stiffness. A three-point flexural test on the honeycomb jamming mechanism and teleoperated object-grasping tasks were conducted to evaluate the device's functionality. Our experiments demonstrated a small RMSE and strong correlations in teleoperated motion, stiffness rendering, and interaction force feedback. The HJ-Haptic effectively adjusts its stiffness in response to real-time gripper feedback, mimicking the sensation of direct object grasping with hands. The device's use of vacuum pressure ensures operator safety by preventing dangerous outcomes in case of gas leakage or material failure. Incorporating the HJ-Haptic into the teleoperation framework provided the reliable perception of object stiffness and stable teleoperation. This study highlights the potential of the honeycomb jamming mechanism for enhancing haptic feedback in various applications, including teleoperation scenarios, as well as interactions with extended-reality environments.

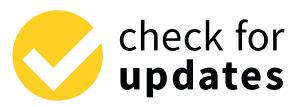

## 1. Introduction

Haptics is essential for bilateral teleoperation [1]. Most systems involve primary and secondary devices [2], where human operators remotely control a secondary device by manipulating the primary device. These systems have been a popular research topic for years due to their extensive applications, such as robot-assisted microsurgery [3], telerehabilitation [4], as well as operations in environments like factories [5], space [6], and underwater [7]. Despite considerable advancements, achieving stable teleoperation with effective haptic feedback remains an open research question.

For intuitive and safe interaction between operators and remote systems [8], the sense of touch, comprising cutaneous and kinesthetic feedback [9], is crucial in tasks like object grasping. Cutaneous feedback, relying on vibrations, indentation, and lateral stretching [10], helps estimate object stiffness. However, it is less intuitive than kinesthetic feedback. Researchers like Pacchierotti [10–12] and Palagi [13] have shown that cutaneous feedback can improve teleoperation stability in kinesthetic devices, yet synchronization issues of cutaneous–kinesthetic feedback from time delays [14,15] may impair operators' telepresence.

Kinesthetic feedback replicates the direct force experienced by the human body [16], making it particularly intuitive for operators to gauge grip force when handling objects of varying stiffness [13]. However, conventional kinesthetic devices faced two major issues. The first is delayed force feedback, which can lead to instability and sensation distortion [17]. Most existing haptic devices for teleoperation rely on admittance and impedance force control, separating stiffness perception into force and motion components [2,18–21]. This separation, coupled with delayed feedback, compromises the reliability of haptic sensations. The second challenge is joint misalignment [22] in hand exoskeletons [23]. Misalignment can result in undesired shear forces on the skin or joint torque, and unreliable motion measurements, degrading the teleoperation experience.

To address these limitations, stiffness-rendering devices like SORI [8] and Hapstick [24] have been explored for their ability to provide reliable stiffness perception directly. However, these devices have yet to be evaluated in teleoperation settings. Our approach is to develop a wearable stiffness-changing haptic device for teleoperated object-grasping tasks. With kinesthetic feedback, this device can render objects' stiffness, enabling remote hand-grasping and grip force control. Worn on the thumb, this device allows unrestricted movement for the other fingers before gripper contact, enhancing the user awareness of object interaction. Inspired by Li's gripper design [25], we integrated a honeycomb jamming mechanism into our device for its lightweight and easily implementable properties. This paper will explore the feasibility of this design in a teleoperation setting.

The rest of the paper is organized as follows. Section 2 describes the honeycomb jamming haptic device (HJ-Haptic) design and its teleoperation framework. Section 3 presents the experimental results of the HJ-Haptic and teleoperation framework in a teleoperated object-grasping task. Section 4 provides a discussion and outlines potential research contributions. Finally, Section 5 concludes the paper.

## 2. Materials and Methods

This paper presents a novel haptic device for teleoperated object-grasping tasks with two significant novelties:

1. The honeycomb jamming mechanism is employed in the haptic device, resulting in a wearable, lightweight (20 g), and compact design that ensure finger mobility. This mechanism allows the device to increase its stiffness up to 2.3 times with a 30 kPa vacuum pressure, offering varying levels of finger kinesthetic feedback in teleoperation.
2. The proposed stiffness-rendering haptic device is integrated into a bilateral teleoperation framework. This device provides real-time kinesthetic feedback during teleoperated object-grasping tasks, allowing operators to feel remote objects' stiffness and gauge their grip force.

### 2.1. Honeycomb Jamming Mechanism

The jamming mechanism shows promise for variable stiffness in haptic devices, as highlighted in Follmer's user interfaces [26] and Stanley's controllable surface [27] with granular jamming, Simon's mitten [28] and Zhang's gloves [29] with layer jamming, and Jadhav's fiber jamming glove [30]. Among these jamming mechanisms, the layer jamming mechanism is more suitable for wearable haptic devices because of its ease of implementation and ability to meet the movement requirements for teleoperation control.

The layer jamming mechanism offers more stable shape control than granular jamming, which can shift to a fluid state when unjammed. This stability simplifies the implementation of haptic devices without additional design efforts to maintain the shape of an unpowered or unjammed device.

As for teleoperating a two-finger gripper in this paper, the haptic device requires only one degree of freedom (DoF) to detect the finger-pressing movement for 1-DoF gripper control (Open/Close) and to provide haptic feedback. Thus, a multi-DoF jamming mechanism like fiber jamming in [24] is unnecessary. Layer jamming offers higher bending stiffness than granular and fiber jamming [31], providing sufficient stiffness for kinesthetic feedback.

Among layer jamming mechanisms, the honeycomb jamming mechanism has many advantages but has received limited attention. Few studies explored its potential in various applications. While previous research on honeycomb jamming in soft grippers [25] had different objectives, we modified the existing honeycomb jamming mechanism for haptic device design.

As detailed in [25], the honeycomb jamming mechanism is based on the layer jamming effect of a sandwich honeycomb structure comprising a honeycomb core and two jamming layers, as shown in Figure 1a. The choice of jamming layer is critical, requiring flexible but not extensible layers. As shown in Figure 2, the flexible material allows the honeycomb structure to remain compliant to external loads (e.g., finger-pressing motion) in the unjammed stage without vacuum pressure. When vacuum pressure is applied, the outer membrane compresses the jamming layers (shown with clamp ③), increasing friction between the honeycomb core and the non-extensible jamming layers ②. This friction constrains relative motion within the structure. Since the jamming layers are non-extensible, the upper and lower layers with constant lengths resist bending motion, causing the structure to become rigid. Hence, the jamming layers should be flexible yet non-extensible. If rigid materials are used, the mechanism would resist finger-pressing motion in both the jammed and unjammed stages. Additionally, if extensible materials are used, the mechanism would bend under finger motion even in the jammed stage, as it would permit relative length differences between the upper and lower layers.

According to [32], the honeycomb core in the sandwich structure can effectively transfer the transverse load into axial loads, meaning the fingertip force applied to the structure is transferred to the non-extensible jamming layers when jammed. The stiffness can be estimated based on the extensibility of the jamming layers. Compared to other layer jamming mechanisms, the honeycomb core reduces the weight and density by replacing most layers with the honeycomb core. Hence, the overall HJ-Haptic is as light as 20.0 g, including the honeycomb jamming mechanism (4.6 g) and other required mechatronic components.

While the working principle is similar to that in [25], significant differences arise in the materials used for the honeycomb core and jamming layers. In [25], a 3D-printed honeycomb core was employed, which may suffer from issues related to brittleness and inconsistent density. In our previous work of jamming gripper [33], we used an aluminum honeycomb core, which is prone to yielding and permanent deformation. Unlike its typical use in composite sandwich in static structure, the honeycomb in haptic devices is subjected to significant dynamic deformation. To address these challenges, we selected a Nomex® aramid paper honeycomb core, which is commercially available and guarantees consistent quality. Its excellent durability and impact resistance make it more suitable for repeated use in a haptic device.

For the jamming layer [25], silicon carbide grit papers were used, but their loose abrasive particles compromised the mechanism's robustness. We chose Velcro strips, commonly used in commercial products, for their securely attached loops, enhancing the jamming effect and overall reliability. Compared to alternatives like paper, Velcro is more durable, resisting wear and tear over time, making it a reliable choice for long-term applications.

Furthermore, the device stiffness reported in [25] was as small as 0.04 N/mm, which is insufficient for haptic devices. Most existing devices ranged from 0.17 to 2.0 N/mm [31,34,35], with some like SORI [8] reaching up to 6.52 N/mm. Thus, we empir-ically modified the honeycomb configuration, including size and material, to achieve a 2.64 N/mm stiffness comparable to other existing devices.

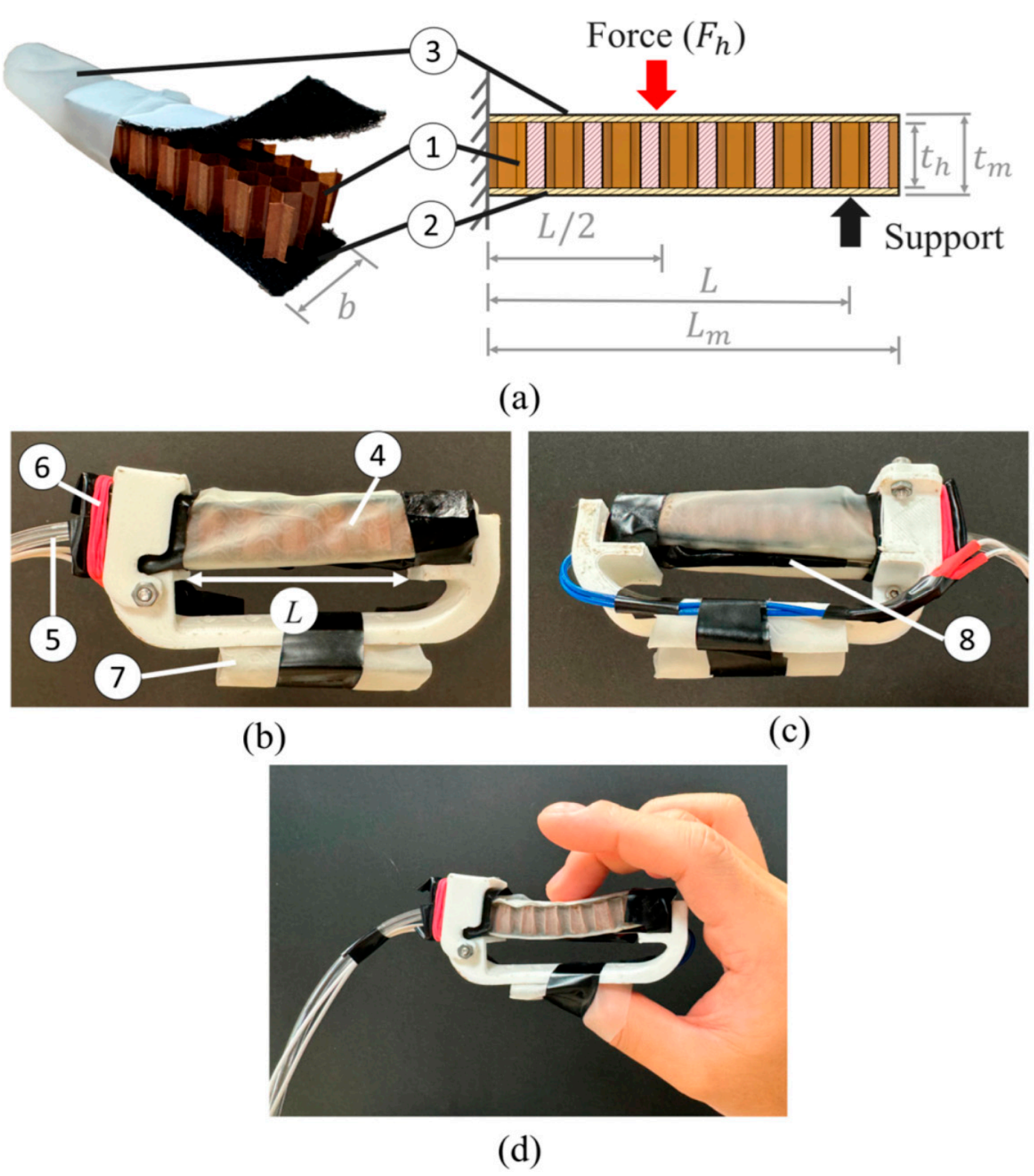

**Figure 1.** HJ-Haptic design. (**a**) The cross-section view of the honeycomb structure and its free body diagram with fingertip force ($F_h$). The structure includes ① the honeycomb core, ② the jamming layers, and ③ the membrane. (**b**,**c**) The side views show ④ the honeycomb structure, ⑤ the silicone tube connected to a vacuum pump, ⑥ the rubber band, ⑦ the thumb attachment, and ⑧ the steel strip with a mounted strain gauge. In (**a**–**c**), $L$ and $L_m$ are the length between the two ends and the structure length from the fixed end; $t_h$ and $t_m$ are the thicknesses of the honeycomb core and whole structure; and $b$ is the width of the structure. (**d**) The HJ-Haptic is worn on the operator's thumb tip.

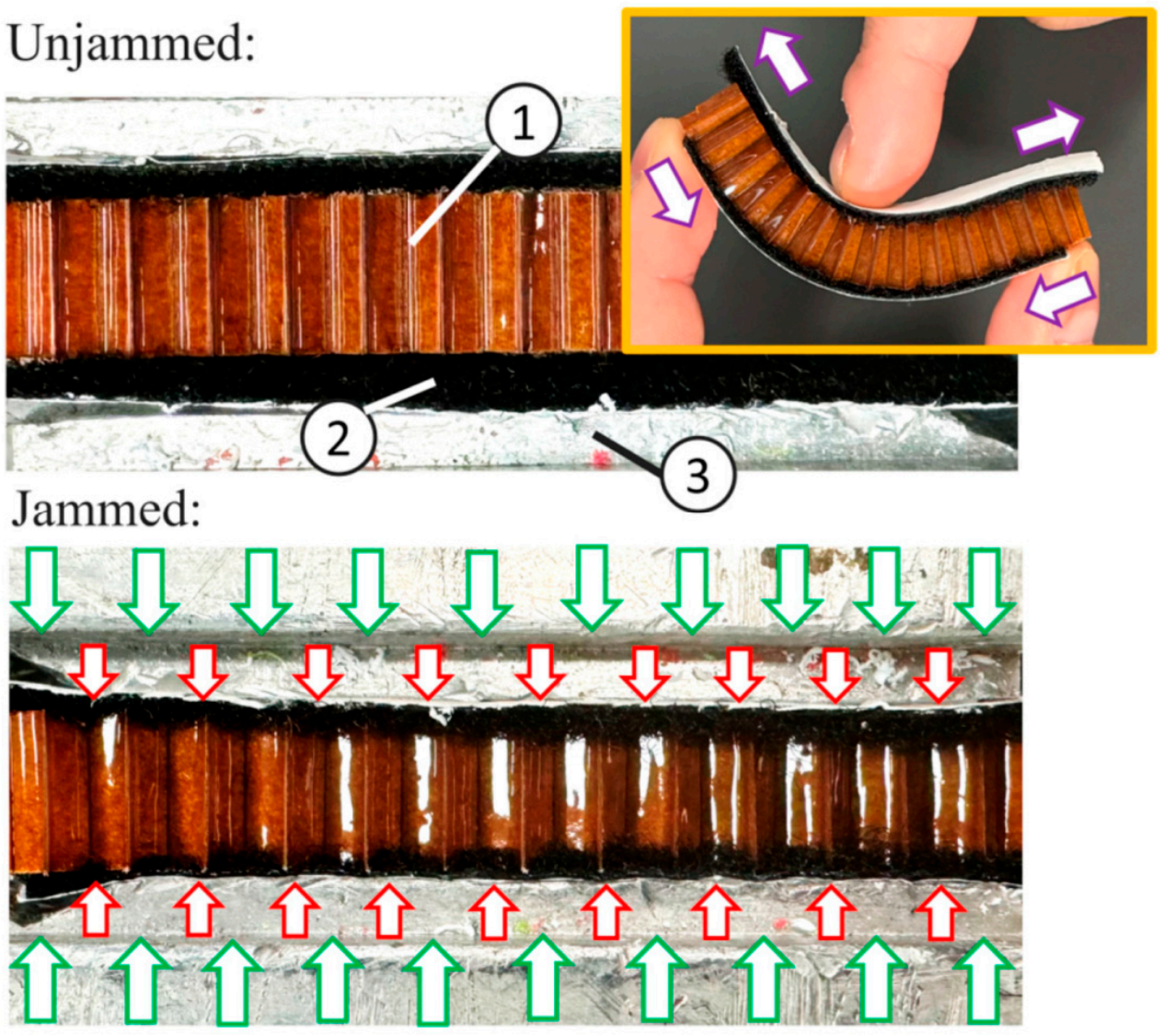

**Figure 2.** Jamming effect of the HJ-Haptic. Figures show ① the honeycomb core, ② the jamming layers, and ③ the clamp that mimics the membrane compression. The purple arrow indicates the sliding direction of the jamming layers, while the green and red arrows represent the applied pressure and resulting force on the jamming layers and honeycomb core, respectively.

*2.2. HJ-Haptic Design*

The HJ-Haptic is lightweight (20 g) and compact (30 mm (width (W)) × 90 mm (length (L)) × 33 mm (thickness (T))), while the honeycomb jamming mechanism only weighs 4.6 g and has a size of 16.6 mm (W) × 73 mm (L) × 16 mm (T). The design allows it to be worn on the thumb without restricting finger flexion, particularly when the operator presses the honeycomb jamming mechanism for a large range of motion up to 10 mm.

As shown in Figure 1a, the key components of the honeycomb jamming mechanism include a honeycomb core ①, two jamming layers ②, and a membrane ③. As mentioned in Section 2.1, our design used a 10 mm-thick commercial grade Nomex® aramid paper honeycomb core with a 4.8 mm cell size. The honeycomb core has dimensions of approximately 16 (W) × 67 mm (L), with a width of 3 cells and a length of 14 cells. The jamming layers are made from the loop side of Velcro strips, as we found that the hooks are too stiff to effectively create friction with the honeycomb core in the jammed stage, which diminishes the jamming effect. For the membrane, we used a finger from a commercial PVC resin glove that is airtight and also elastic and thick (0.2 mm) enough to conform to the honeycomb structure and compress the jamming layers when the vacuum is applied. Additionally, other parameters of honeycomb structure can be found in Table 1.

**Table 1.** Parameters of honeycomb jamming mechanism.

| Parameters | $L$ | $L_m$ | $t_h$ | $t_m$ | $b$ |
|---|---|---|---|---|---|
| Value (mm) | 53.5 | 63.5 | 10.0 | 13.0 | 17.0 |

$L$ and $L_m$ are the length between the two ends and the structure length from the fixed end; $t_h$ and $t_m$ are the thicknesses of the honeycomb core and the whole structure; and $b$ is the width of the structure.

To implement this honeycomb jamming mechanism into a haptic device, we treated it as a beam that was fixed at one end and supported at the other with a single point load (i.e., fingertip force ($F_h$)), as shown in Figure 1a. In Figure 1b, the device structure is 3D-printed using polylactic acid plastic (PLA). The honeycomb structure ④ is connected to a vacuum pump via a silicone tube ⑤, with a rubber band ⑥ used to prevent gas leakage. The operator can wear the haptic device with the thumb attachment ⑦, as shown in Figure 1d.

Additionally, we installed a 4 mm (W) × 70 mm (L) × 0.25 mm (T) steel strip ⑧ to provide the elastic effect, allowing the honeycomb jamming mechanism to return to its non-deflected state after finger pressing. In this proof-of-concept paper, we intentionally selected this dimension to achieve a lower device stiffness, which is comparable to other devices (0.17 to 2 N/mm [31,34,35]). However, it should be stiff enough for the teleoperated object-grasping task. Nevertheless, this stiffness can be easily increased for other applications using a thicker or wider steel strip.

As mentioned in Section 2.1, vacuum pressure controls the stiffness of the honeycomb jamming mechanism. We used a DC pump to adjust vacuum pressure using PID control for stiffness control. The voltage control command ($V$) to vacuum-pump is described using the following equations:

$$e(t) = p_d(t) - p_h(t) \tag{1}$$

$$V(t) = K_p e(t) + K_i \int_0^t e(\tau) d\tau + K_d \left( \frac{d}{dt} e(t) \right) \tag{2}$$

where $p_d$ and $p_h$ are the desired pressure and pressure feedback from a pressure sensor, respectively, and $K_p,\ K_i,\ K_d$ are the PID control gains that were tuned empirically.

As for sensing, we implemented a pressure sensor to measure the vacuum pressure of the honeycomb structure as feedback for the PID controller. Additionally, as shown in Figure 1c, we used a strain gauge mounted on the steel strip ⑧ to measure the deflection of the honeycomb structure due to fingertip force ($F_h$).

Regarding wearability, Figure 1d shows that the HJ-Haptic can be worn and secured to the thumb's distal phalanges with a strap. Given that the strap does not have direct contact with other fingers, the HJ-Haptic allows free flexion motion of other fingers during finger pressing.

### 2.3. Stiffness-Rendering Teleoperation Framework

Beyond device design, it is essential to demonstrate its functionality within the teleoperation framework. The existing literature shows limited teleoperation frameworks that render object stiffness as haptic feedback, enabling operators to adjust grip force based on finger sensation.

Current bilateral teleoperation frameworks, including admittance and impedance force control, separate stiffness into force and motion components [2,18–21]. In these systems, operators control the robot using motion or force command while receiving force feedback as kinesthetic feedback. However, due to inherent telecommunication delays [14,15], many bilateral teleoperation systems struggle to synchronize these components, making it challenging to accurately render object stiffness and interaction behavior. This lack of synchronization often results in issues with robot accuracy and stability, complicating teleoperated tasks like remote object grasping.

For instance, in an impedance control framework, an operator manipulates the gripper motion remotely. After a delay, the gripper grasps the object and sends the measured interaction force feedback to the haptic device. Another time delay occurs before the haptic device provides kinesthetic feedback to the operator. Consequently, the operator's finger motion and kinesthetic feedback may not be synchronized, leading to a false perception of object stiffness. Ideally, the system should measure, transmit, and render object stiffness directly in one component to ensure synchronization, even with time delays.

To address this, we implemented a stiffness-rendering bilateral teleoperation framework, as shown in Figure 3. The framework begins when the operator applies a fingertip force ($F_h$), and the strain gauge ($\Delta x_h$) measures the deflection of the honeycomb structure. This measurement is used to command the gripper's fingertip to move horizontally. Using Equation (3), we computed the desired angle command ($q_d$) for the gripper's motor to match $\Delta x_h$. Simultaneously, the load cell of the gripper measures real-time force feedback ($F_g$) as the gripper presses the objects. Using Equation (4), we estimated the object stiffness ($k_o$). This estimated stiffness was then mapped to the desired pressure ($p_d$) for the vacuum pump using Equation (8). Then, we controlled the HJ-Haptic's stiffness through PID pressure control using Equation (2).

$$q_d(\Delta x_h) = \cos^{-1}\left(1 - \frac{\Delta x_h}{L_g}\right) \tag{3}$$

$$k_o\left(\Delta x_h, F_g\right) = \frac{F_g}{\Delta x} \tag{4}$$

where $L_g$ is the length of links in a four-bar mechanism of the gripper.

In this paper, we used a widely adopted two-finger parallel gripper to test our teleoperation framework, as shown in Figure 4a. The gripper utilizes a four-bar mechanism with four identical links ($L_g = 50$ mm). This mechanism allows the gripper tip to move horizontally to match $\Delta x_h$. As a result, it simplifies the calculation of object stiffness estimation by eliminating concerns about the orientation of the gripper fingertip.

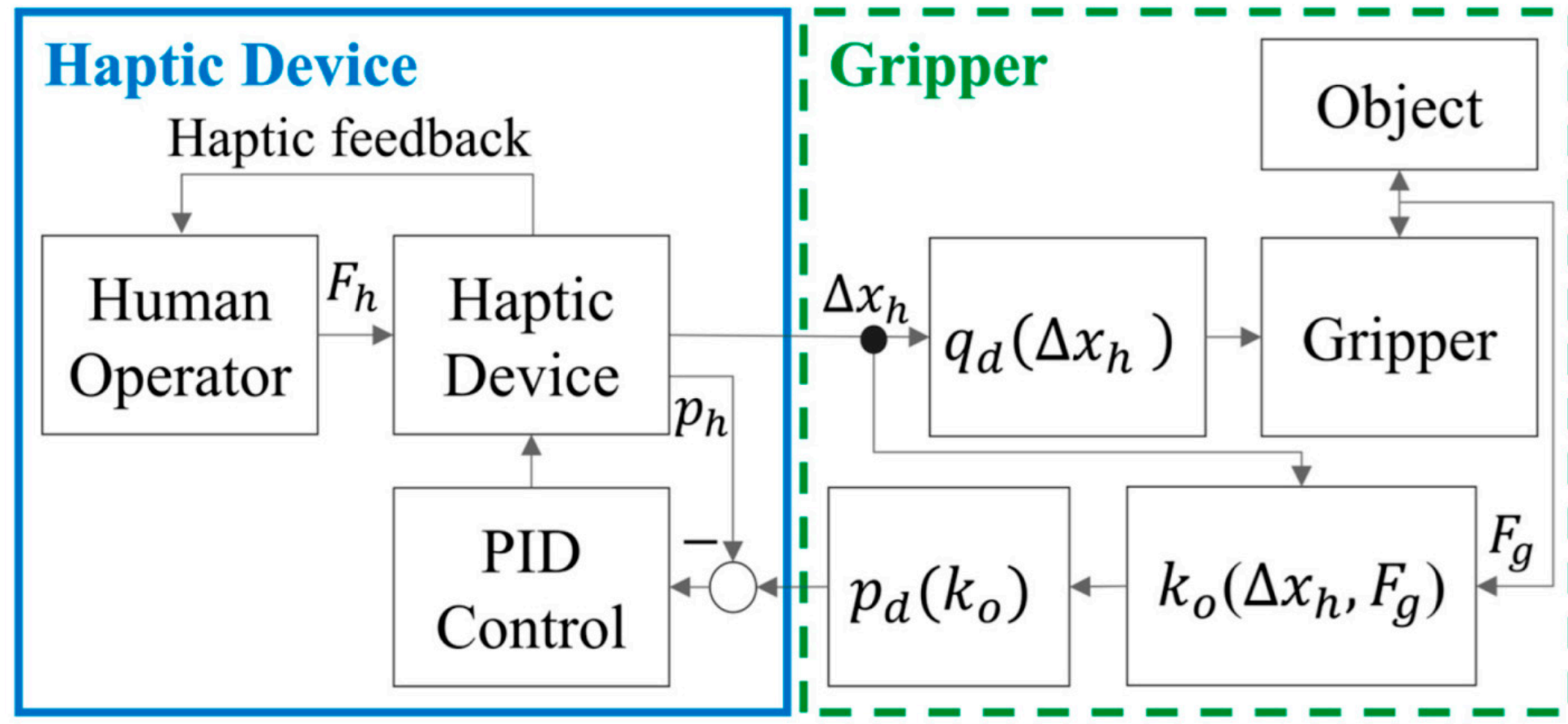

**Figure 3.** The stiffness-rendering teleoperation framework for the HJ-Haptic. $F_h$ and $F_g$ represent fingertip force and gripper force feedback. $\Delta x_h$ indicates the changes in strain gauge measurement. $q_d$ and $k_o$ are the desired angle commands for gripper's motor (Equation (3)) and estimated object stiffness (Equation (4)). $p_d$ and $p_h$ are the desired pressure and pressure feedback from a pressure sensor, respectively.

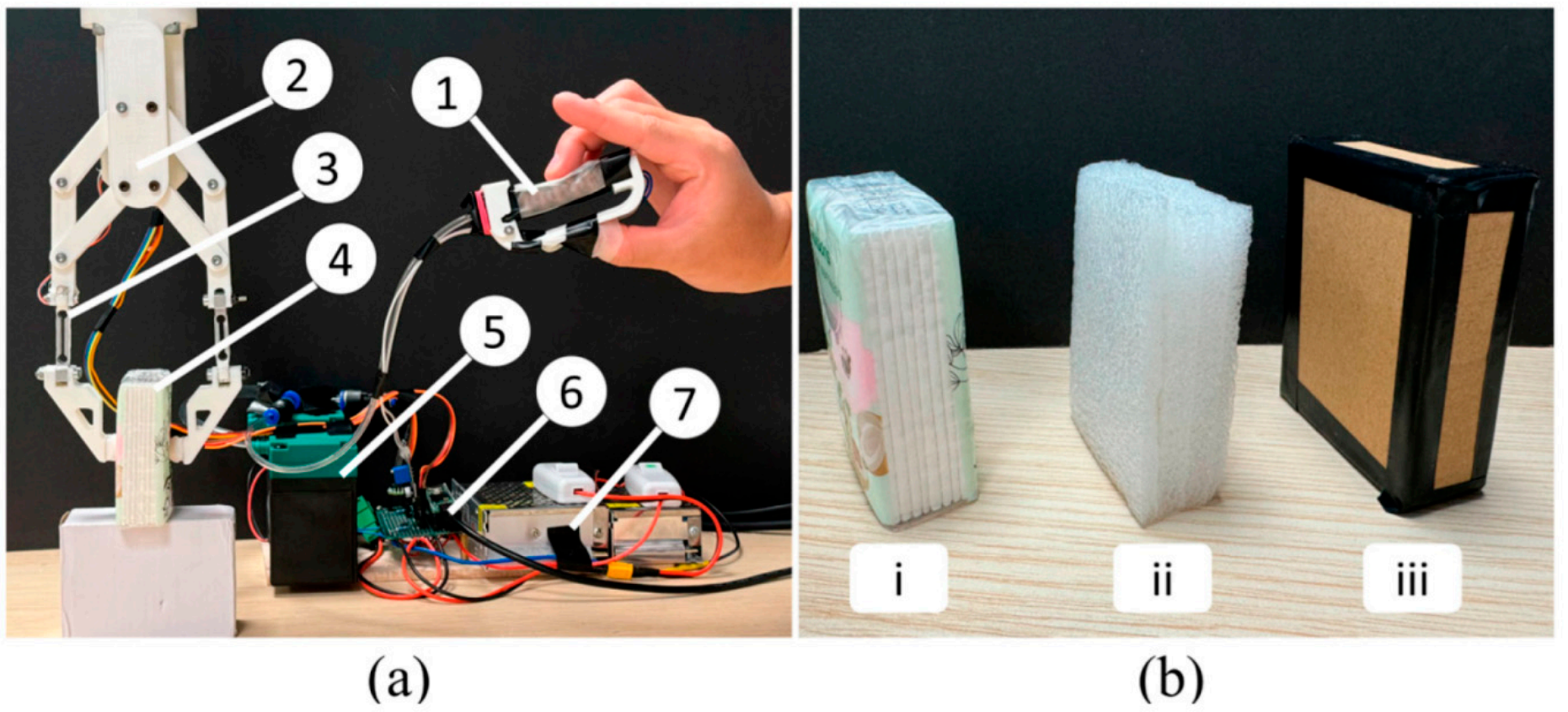

**Figure 4.** (**a**) Experimental evaluation of haptic device during gripper teleoperation. The equipment includes ① the HJ-Haptic, ② the gripper, ③ the load cell, ④ the HJ-Haptic, ⑤ the vacuum pump, ⑥ the microcontroller, and ⑦ the power supply. (**b**) Three objects with different stiffness levels: (**i**) tissue paper, (**ii**) foam sponge, and (**iii**) carton box.

## 3. Results

### *3.1. Device Characterization*

#### 3.1.1. System Hardware

We implemented the HJ-Haptic, remote gripper, and real-time teleoperation framework into a microcontroller Teensy 4.1. For the HJ-Haptic, the vacuum pressure is provided by a DC pump with a maximum pressure of 90 kPa. When the Teensy 4.1 receives the pressure command, it sends pulse-width modulation (PWM) signals to an N-Channel MOSFET (RFP30N06LE). It controls the DC pump via voltage input. A pressure sensor (XGZP6847A) provides pressure feedback for PID control. Additionally, we integrated a quarter-bridge strain gauge into Teensy 4.1 to measure the deflection of the honeycomb structure caused by the operator's finger movement.

The remote gripper, shown in Figure 4a, uses a Digital Servo (DS3235) to actuate the two-finger parallel gripper with a four-bar mechanism. The motor angle is sent from the Teensy 4.1 via PWM signals. For object stiffness estimation, we measured the force feedback using a load cell connected to the Teensy 4.1 via an HX711 amplifier.

### 3.1.2. Effects of Pressure

To study the stiffness-changing behavior of the HJ-Haptic, we performed a three-point flexural test. The experiment setup is shown in Figure 5a. Initially, we recorded a 5 s baseline for the strain gauge ⑤. We then placed the load ① on the dial meter ②. The spherical indenter ③ of the dial meter, with a diameter of 15 mm, simulated a fingertip by applying a single-point load to the HJ-Haptic ④. After about 5 s, the dial meter and strain gauge measurements stabilized. We recorded data for 10 s for analysis. We conducted five trials for each condition, with pressure ranging from 0 to 30 kPa and load ranging from 200 g to 1000 g.

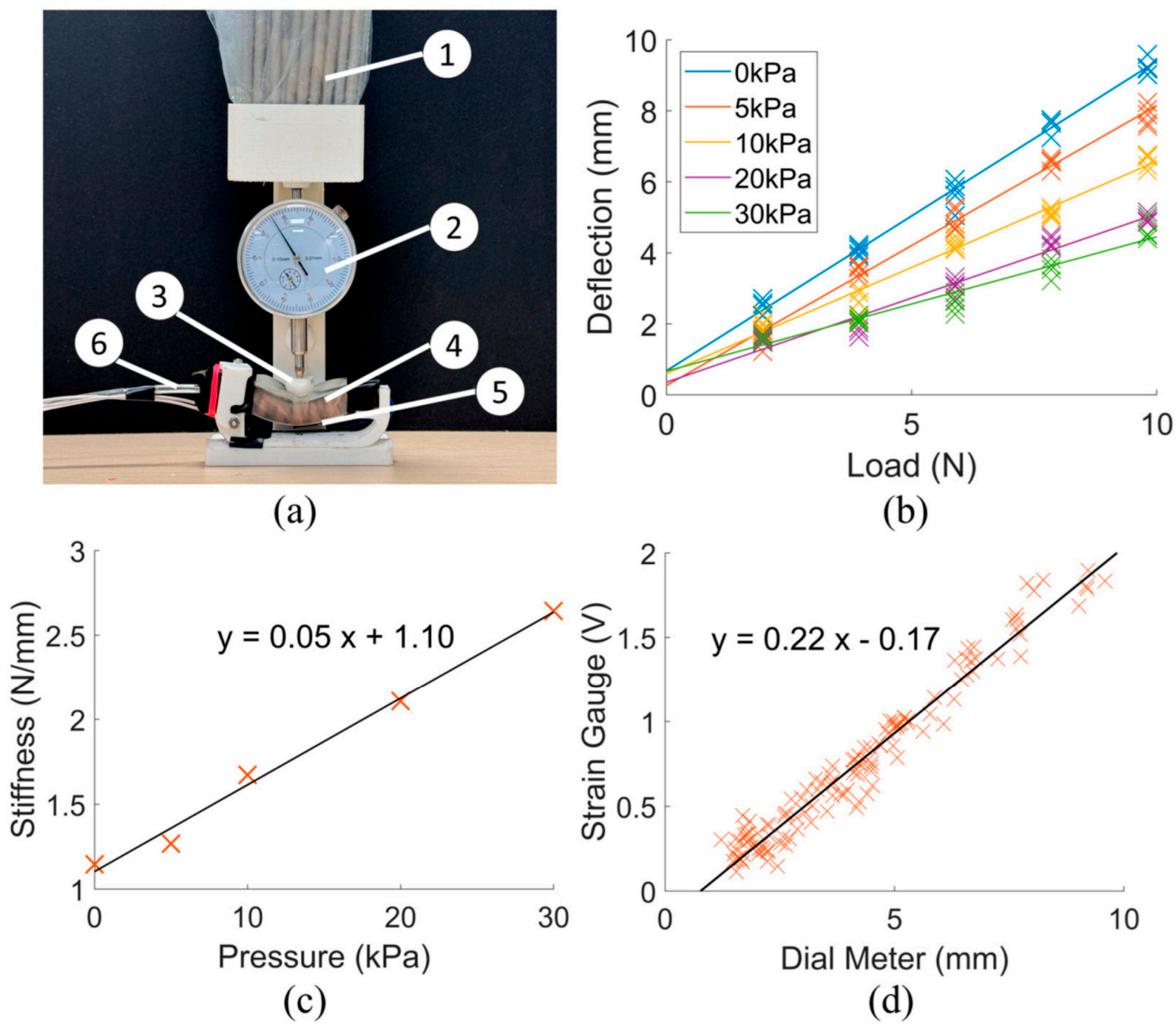

**Figure 5.** (**a**) Experiment setup for the three-point flexural test. The equipment includes ① the load, ② the dial meter, ③ the indenter, ④ the HJ-Haptic, ⑤ the strain gauge, and ⑥ the tube. (**b**) The effect of vacuum pressure on honeycomb structure stiffness, ranging from 0 kPa to 30 kPa. (**c**) The linear relationship between the stiffness and vacuum pressure of the honeycomb structure. (**d**) Linear relationship between the dial meter and strain gauge.

The experimental results are shown in Figure 5b, where each data point represents an average of five trials. The results indicate that the stiffness of the HJ-Haptic increases with a higher vacuum pressure. We applied linear regression to determine the stiffness of the honeycomb structure ($F/\delta$) for each pressure condition, using the average load ($F$) and deflection ($\delta$). The $R^2$ values of the regressions ranged from 0.941 to 0.990. Figure 5c and Table 2 show the relationship between pressure and stiffness, achieving stiffness changes ranging from 1.15 to 2.64 N/mm with a 30 kPa vacuum pressure. Thus, the HJ-Haptic can achieve a higher stiffness-changing ratio per kPa of 0.077 compared to the other existing pneumatic stiffness-varying haptic devices outlined in Table 3.

**Table 2.** The relationship between vacuum pressure, stiffness, and flexural modulus.

| Vacuum Pressure (kPa) | 0 | 5 | 10 | 20 | 30 |
|---|---|---|---|---|---|
| Stiffness (N/mm) | 1.15 | 1.27 | 1.67 | 2.11 | 2.64 |
| Flexural modulus (MPa) | 0.51 | 0.57 | 0.75 | 0.95 | 1.19 |

**Table 3.** A comparison with existing pneumatic stiffness-varying haptic devices.

| | Type [1] (K/C) | Jamming [2] (L/F) | Wearable (Y/N) | Stiffness (N/mm) [Pressure (kPa)] | Ratio per kPa [3] |
|---|---|---|---|---|---|
| HJ-Haptic | K | L | Y | 1.15–2.64 [0–30] | 0.077 |
| Zhang's gloves [24] | K | L | Y | 0.15–0.47 [4] [0–60] | 0.052 |
| Hapstick [19] | K | F | N | 38.6–192.6 [0–70] | 0.071 |
| SORI [8] | K, C | / | N | 0.16–3.62 [5,6] [0–200] | 0.11 [6] |
| Tactile displays [28] | C | / | Y | 0.075–0.17 [4] [5–20] | 0.15 |

[1] (K) kinesthetic and (C) cutaneous; [2] jamming: (L) layer and (F) fiber; [3] stiffness-changing ratio per unit of pressure (kPa); [4] estimated from figures; [5] estimated by a linear model; [6] estimated for the kinesthetic part.

Additionally, the flexural modulus of the HJ-Haptic at different vacuum pressure levels can be computed using the following equations:

$$E = \frac{7L^3}{768I}\left(\frac{F}{\delta}\right) \quad (5)$$

$$I = \frac{bt_m{}^3}{12} \quad (6)$$

where $L$ is the length of the honeycomb structure between two ends shown in Figure 1a,b, and $t_m$ and $b$ are the thickness and width of the structure, respectively. The parameters of honeycomb structure and the computed values are shown in Tables 1 and 2.

#### 3.1.3. Stiffness Characterization

In Figure 5c, a linear relationship between stiffness ($k_h$) and pressure ($p_h$) is observed. By applying linear regression to model this relationship, we derived the following equation, with an $R^2$ value of 0.991:

$$k_h(p_h) = 0.05\ p_h + 1.10 \quad (7)$$

As shown in the teleoperation framework (Figure 3), we can adjust HJ-Haptic stiffness using vacuum pressure. Specifically, after estimating object stiffness ($k_o$), we can apply Equation (8) to determine the desired pressure ($p_d$).

$$p_d(k_o) = 20.0\ (k_o - 1.10) \quad (8)$$

#### 3.1.4. Deflection Measurement with Strain Gauge

We used a strain gauge instead of a dial meter in the teleoperation framework for easier implementation. As shown in Figure 5d, there is a strong Pearson correlation ($r = 0.977, p < 0.001$) between the measurements of the dial meter (in mm) and strain gauge (in V). With these measurements, we found a linear relationship $y = 0.22x - 0.17$ with $R^2 = 0.954$ that was used in our strain gauge calibration.

### *3.2. Experiment with the Teleoperated Grasping Task*

To evaluate the functionality of the HJ-Haptic in teleoperation, we conducted an experiment with a teleoperated object-grasping task. The experiment setup and the human grasping motion are shown in Figures 4a and 6. We started with a 5 s preparation period to record a baseline for the strain gauge. The operator then pressed the HJ-Haptic ① with his middle finger. During the motion adjusting period, the remote gripper ② fine-tuned its finger position to touch the object ④ based on the human input. Upon contact, the load cell ③ recorded its measurement as the initial point, accounting for any potential influence from the gripper's components. Simultaneously, the gripper recorded its position as the initial point and began estimating the object's stiffness using the position and force measurement relative to their initial points.

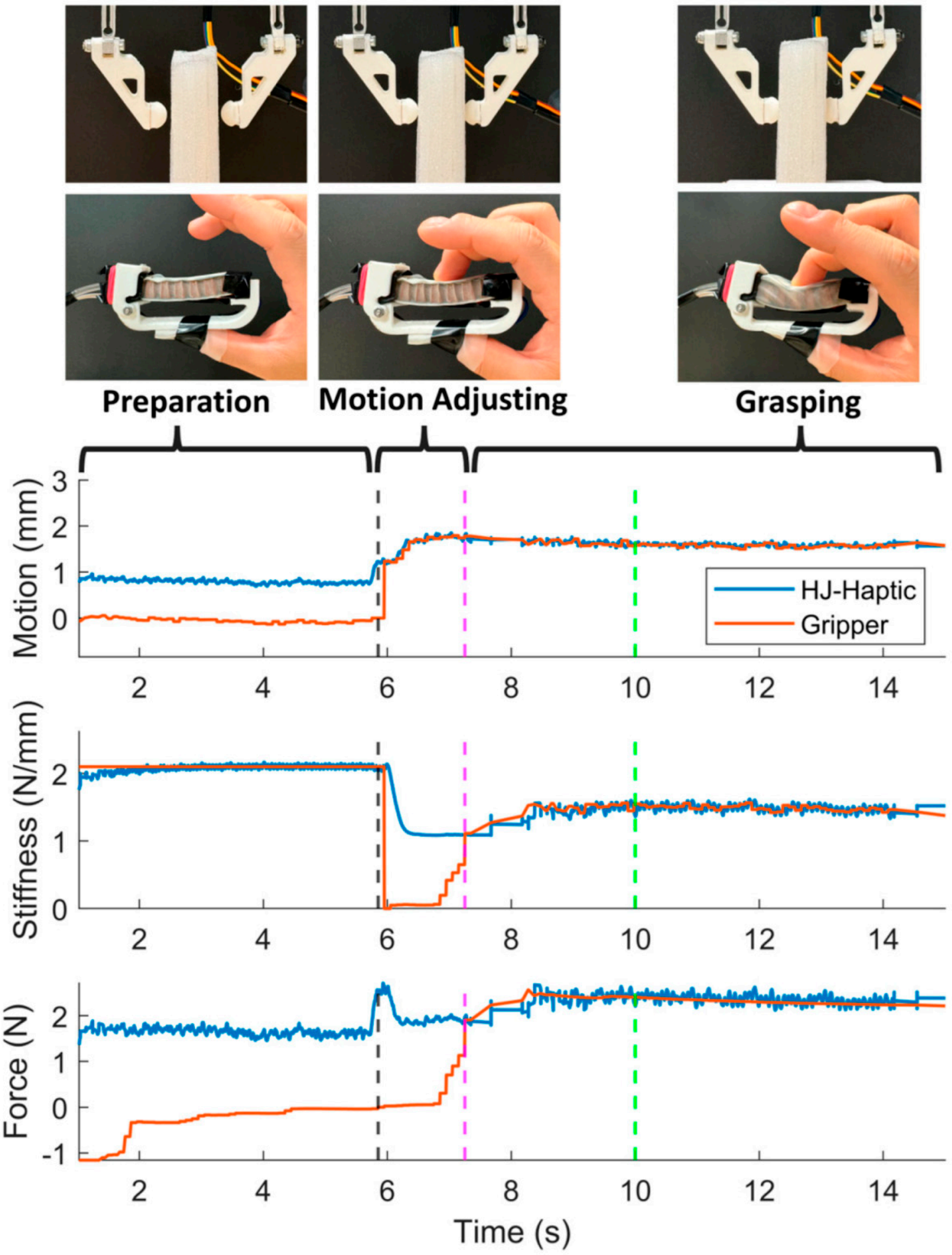

**Figure 6.** The object grasping motion in the teleoperated task. The black and purple dotted lines indicate the onset and end of motion adjusting period, while the green dotted line represents the onset of the stabilized grasping period.

Given that the minimum stiffness of the HJ-Haptic is 1.15 N/mm, we set this as the threshold to initiate the grasping period. During this period, the HJ-Haptic rendered the stiffness to match the estimated object stiffness. The operator could sense the kinesthetic feedback and apply the appropriate grip force for gripper control.

For this experiment, we conducted five trials for each object. The objects with different stiffness levels were tissue paper, foam sponge, and carton box, as shown in Figure 4b.

3.2.1. Teleoperation Evaluation

To evaluate the teleoperation performance, we compared three pairs of data from the HJ-Haptic and remote gripper during the whole grasping period: (i) motion with strain gauge measurement ($\Delta x_h$) and fingertip displacement of the gripper, (ii) stiffness with the HJ-Haptic stiffness ($k_h$) and object stiffness estimation ($k_o$), and (iii) interaction force with fingertip force ($F_h = k_h \Delta x_h$) and load cell measurement ($F_g$).

As shown in Figure 7, a small root-mean-square error (RMSE) and weak–strong correlations ($p < 0.001$) were observed for motion, stiffness, and interaction force across three different objects. These results indicate that the mentioned data pairs from the HJ-Haptic and remote gripper were closely aligned and correlated. This suggests that the HJ-Haptic uses operators' finger motion and fingertip force to command the remote gripper for object grasping. Additionally, the HJ-Haptic adjusts its stiffness to match the estimated object stiffness reliably.

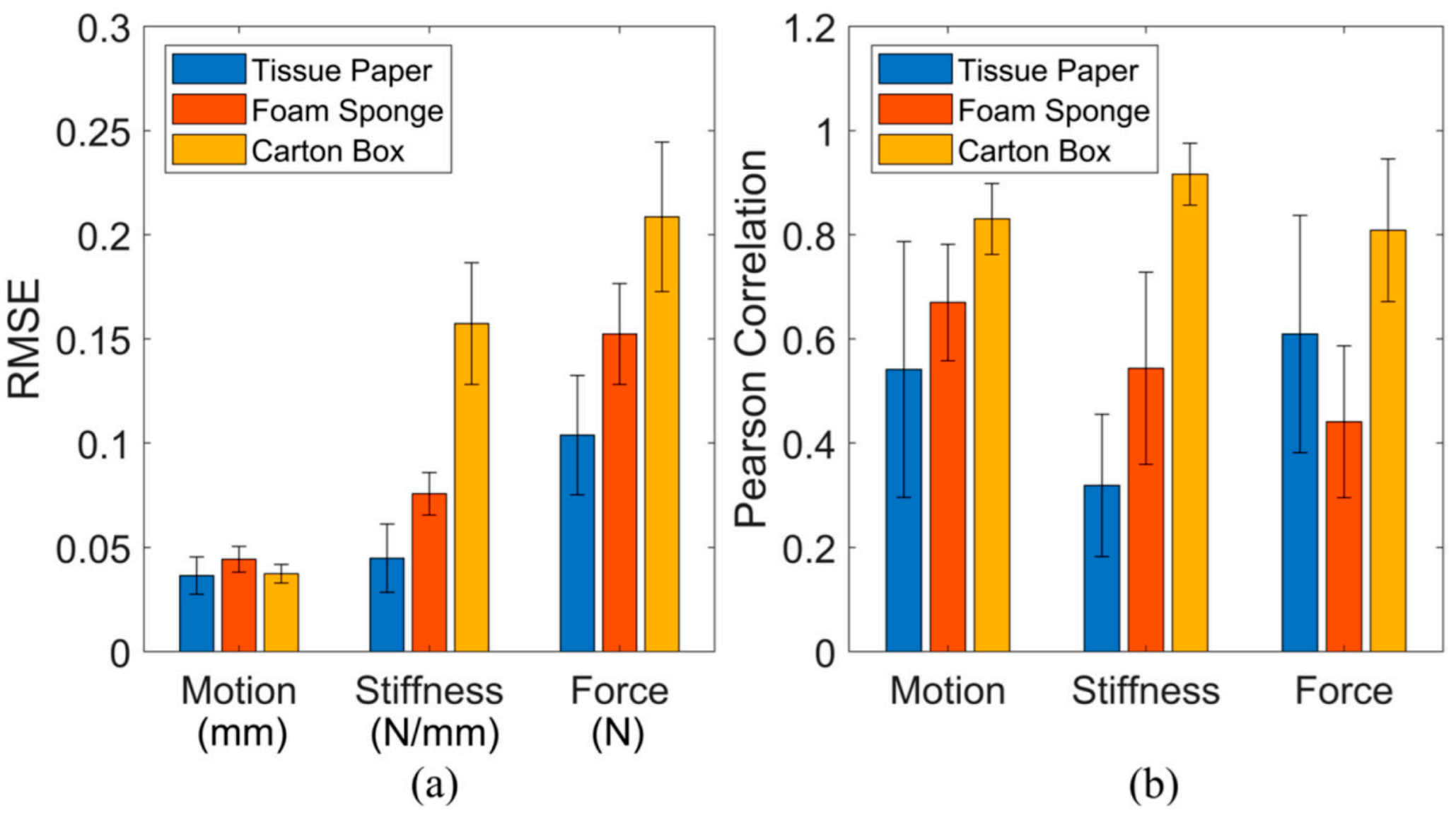

**Figure 7.** The (**a**) root-mean-square errors and (**b**) Pearson correlations (mean ± SD) of motion, stiffness, and force between the HJ-Haptic and gripper in 5 trials for each object.

However, the RMSE of stiffness and interaction force increased when rendering a more rigid object but remained small. One possible explanation is gas leakage at a higher vacuum pressure, which can compromise the system's ability to achieve the desired stiffness. Another reason may be the PID control error. The control gains were intentionally kept small to ensure system stability, which may have limited the controller's robustness. As a result, the system may struggle to accurately track higher vacuum pressures, leading to larger RMSE values.

The Pearson correlation of stiffness was weaker in the soft object. This is because the HJ-Haptic reached its limit ($k_h \leq 1.15$), which can be observed in Figure 8a when grasping the tissue paper. Nevertheless, in Table 4, the overall results across different trials show a small RMSE and strong correlations ($p < 0.001$) in motion, stiffness, and interaction force.

**Table 4.** RMSE and Pearson correlation of motion, stiffness, and force in all trials (*Mean* ± *SD*).

| | **Motion** | **Stiffness** | **Force** |
|---|---|---|---|
| RMSE | 0.04 ± 0.007 mm | 0.09 ± 0.05 N/mm | 0.15 ± 0.05 N |
| Correlation ($p < 0.001$) | 0.68 ± 0.19 | 0.64 ± 0.28 | 0.62 ± 0.22 |

Hence, we showed the possibility of using the HJ-Haptic to control the gripper and render the object stiffness within this teleoperation framework. Also, the gripper can apply a similar grip force on the object as the operator does on the HJ-Haptic.

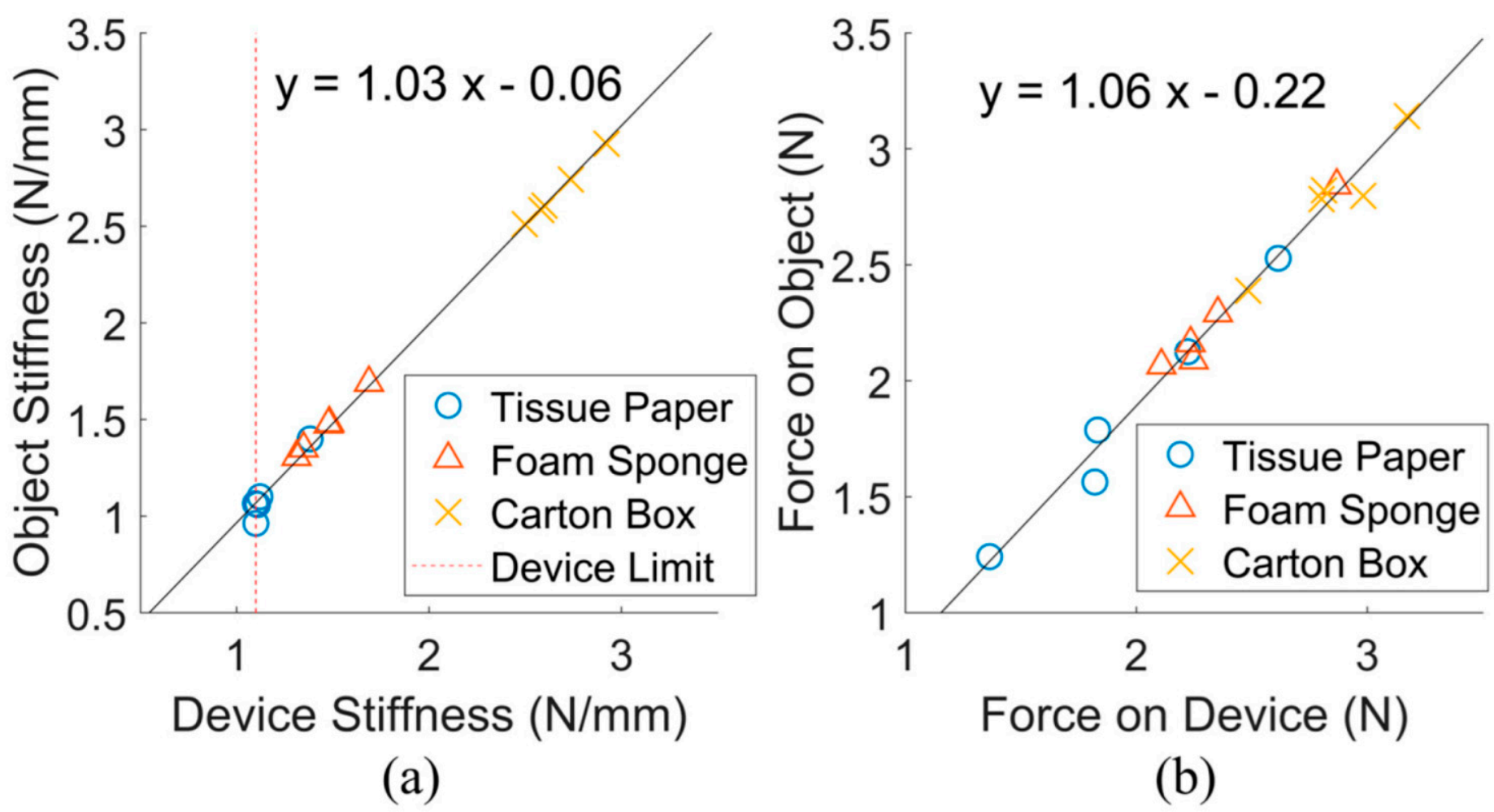

**Figure 8.** (**a**) The stiffness relationship between objects and the device. (**b**) The force relationship between the gripper and the operator.

### 3.2.2. Object Stiffness Evaluation

This section evaluates the HJ-Haptic's stiffness-rendering capability and grip force control. We averaged the stiffness and force feedback in the stabilized grasping period (i.e., 10–15 s), as stated in Figure 6, and compared the feedback of the HJ-Haptic and the remote gripper.

The results in Figure 8a show a linear relationship between the stiffness of objects and the device, with a strong Spearman correlation ($r = 0.989$, $p < 0.001$). The result linear equation is $y = 1.03x - 0.06$ with $R^2 = 0.998$. The slope of 1.03 being close to 1 indicates that the stiffness difference between objects and the device is minimal. The resulting RMSE is 0.040 N/mm. In this figure, the three different object stiffness levels can be distinguished in different clusters. As expected, the carton box is stiffer than the foam sponge, which is stiffer than the tissue paper. This indicates that the haptic device can effectively render the object's stiffness using real-time stiffness feedback from the remote gripper.

In Figure 8b, a linear relationship is also observed between the fingertip force applied by the gripper and the operator, showing a strong Spearman correlation ($r = 0.979$, $p < 0.001$). The resulting linear equation is $y = 1.06x - 0.22$ with $R^2 = 0.985$. The slope of 1.06 being close to 1 indicates that the force difference applied by the gripper and the operator is minimal. The RMSE is 0.11 N. This demonstrates that operators can control the gripper's grip force during teleoperation, which is particularly useful when applying and optimizing a suitable force to manipulate soft objects or objects with unknown stiffness.

## 4. Discussion

The honeycomb jamming mechanism can be effectively used in a stiffness-rendering haptic device, and the HJ-Haptic shows promising results in teleoperation. Section 3.1 demonstrates that the honeycomb jamming mechanism in the HJ-Haptic can change stiffness, ranging from 1.15 N/mm to 2.64 N/mm with a vacuum pressure of 30 kPa. It achieves a relatively high stiffness (2.64 N/mm) with just 30 kPa of vacuum pressure, consuming less energy than some existing haptic devices like SORI [8], which provides only 0.68 N/mm at 30 kPa pressure.

Additionally, we showed the potential of teleoperation with a small RMSE and strong correlations in teleoperated object-grasping experiments. The HJ-Haptic can adjust its stiffness according to the estimated object stiffness from the remote gripper feedback, allowing the operator to feel remote objects' stiffness and gauge their grip force.

The HJ-Haptic directly renders object stiffness based on gripper feedback without separating it into force and motion components, in contrast to existing frameworks [2,24–26]. This stiffness-rendering approach (Figure 9a) facilitates intrinsic signal synchronization with the perceived object stiffness, using the following equations:

$$k_o(t+2\Delta t) = \frac{F_g(t+2\Delta t)}{\Delta x_h(t+2\Delta t)} \tag{9}$$

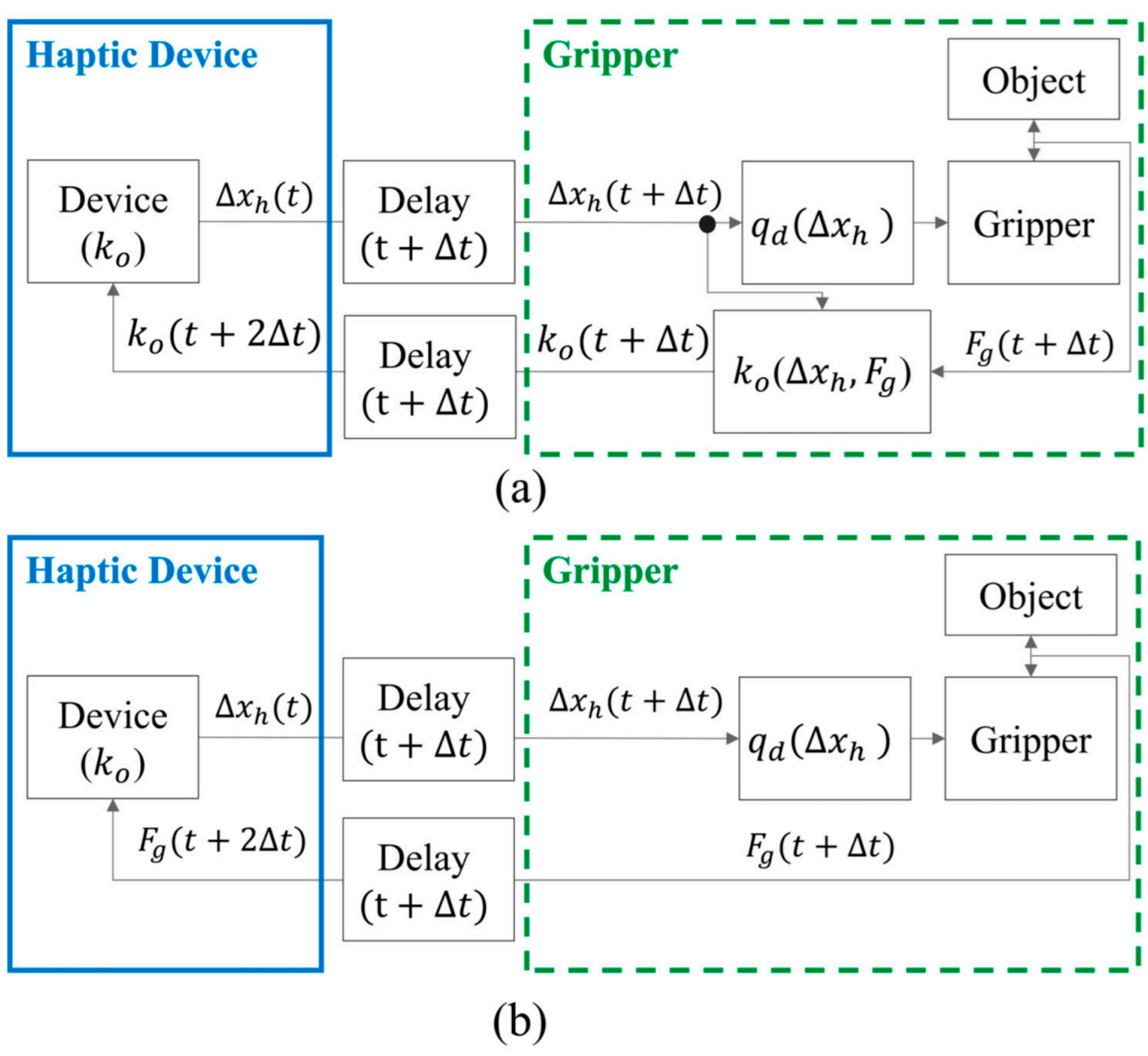

**Figure 9.** Framework for the delay evaluation, where $\Delta t$ represents the delay. (**a**) Stiffness-rendering framework; (**b**) force-rendering framework. $F_g$ represents the gripper force feedback. $\Delta x_h$ indicates the changes in strain gauge measurement. $q_d$ and $k_o$ are the desired angle commands for the gripper's motor and perceived object stiffness.

In comparison, the perceived stiffness in the force-rendering approach (Figure 9b) is as follows:

$$k_o(t+2\Delta t) = \frac{F_g(t+2\Delta t)}{\Delta x_h(t)} \tag{10}$$

While the stiffness-rendering approach may still experience a time delay in rendering stiffness, it theoretically offers reliable haptic feedback without the signal asynchronization issues associated with the force-rendering approach [17]. Although we have not conducted experiments with significant telecommunication delays, our findings in Section 3.2.2, which demonstrate a small RMSE and strong correlations between the device and object stiffness in teleoperation settings, suggest that the stiffness-rendering HJ-Haptic can deliver a reliable haptic sensation for object grasping and enable stable bilateral teleoperation. As part of future work, we plan to conduct experiments under significant telecommunication delays to validate this hypothesis. Furthermore, we anticipate that advancements in 5G network technology will further reduce these time delays, enhancing the overall performance of the system.

When providing kinesthetic feedback, the HJ-Haptic avoids the joint misalignment issues common in hand exoskeleton [23], simplifying finger motion measurements and fingertip force feedback. Instead, the HJ-Haptic uses a strain gauge and honeycomb jamming mechanism to reliably measure fingertip movements and force, ensuring a smooth haptic and teleoperation experience.

Compared to existing haptic devices with stiffness-rendering mechanisms like SORI [8], the HJ-Haptic offers a viable implementation in teleoperation framework with stiffness feedback and rendering. Additionally, its wearable, lightweight, and compact design allows for greater flexibility in teleoperation scenarios, unlike other existing haptic devices [24,35] that are grounded and less portable.

Another advantage of the honeycomb jamming mechanism is its safety. In contrast to some haptic devices that utilize positive pneumatic pressure [8,34], the HJ-Haptic relies on vacuum pressure. In the event of gas leakage or component failure, the device remains secure, ensuring operator safety. This is particularly important for a device that directly interacts with humans, where safety considerations are critical. During our experiments, we observed occasional gas leakage, especially under high vacuum pressure, but no significant safety risks were noted. The HJ-Haptic gradually released vacuum pressure rather than causing any abrupt failure. Nonetheless, we plan to conduct more comprehensive safety tests under high vacuum pressure to further ensure user safety in future applications.

Regarding the ergonomics of the thumb-mounted device, the HJ-Haptic is designed to be worn on the thumb tip. Weighing only 20 g, it places minimal burden on the thumb. The strap only secures the device to the distal phalanges while allowing movement starting from the interphalangeal joint, enabling free thumb motion. When operators perform a grasping motion with the thumb and another finger, similar to a two-fingered gripper, the compact design of the HJ-Haptic facilitates the free movement of the other fingers. The finger mobility is shown in Figure 10. Although we did not observe adverse effects during experiment with one of the authors, we are yet to conduct user studies to assess ergonomics of the device over extended periods.

This paper also highlights several research opportunities. While cutaneous feedback is not the primary aim of this paper, the HJ-Haptic provides some cutaneous feedback in terms of indentation depth and changing contact area when the HJ-Haptic deforms and increases the contact area with the finger (Figure 6). These features may enhance stiffness perception for human operators [8,36]. However, if the honeycomb jamming mechanism proves insufficient for effective cutaneous feedback, we may explore integrating additional mechanisms, as demonstrated in [8,10].

Another research direction is to increase the stiffness-changing ratio. Table 3 highlights the limited ratio of honeycomb jamming mechanism compared to mechanisms like the toroidal soft pneumatic actuator [8] and elastomeric membrane [34]. But simulations in [25] suggested that the honeycomb jamming mechanism could achieve a ratio of up to 14 with specific configurations, though this was not experimentally validated. In our study, while we observed a linear stiffness–pressure relationship, we could not fully explore its potential due to gas leakage in the HJ-Haptic beyond 35 kPa. At higher pressures, increased friction between jamming layers could introduce nonlinear behavior [37], limiting the stiffness–changing ratio. Future work should optimize parameters such as cell size and wall thickness to enhance the ratio and evaluate performance at higher vacuum pressures.

This paper demonstrates the potential of the HJ-Haptic in robotic teleoperation. Our experiments focused on objects with constant stiffness. In future work, we plan to evaluate the device's performance with objects that exhibit non-uniform stiffness. We believe that the HJ-Haptic can effectively render this stiffness, provided that the robot gripper delivers accurate measurements of force and motion. If the gripper can estimate non-uniform stiffness, the HJ-Haptic should also be capable of rendering it, given the strong correlation observed between object stiffness and device stiffness in Figure 8a.

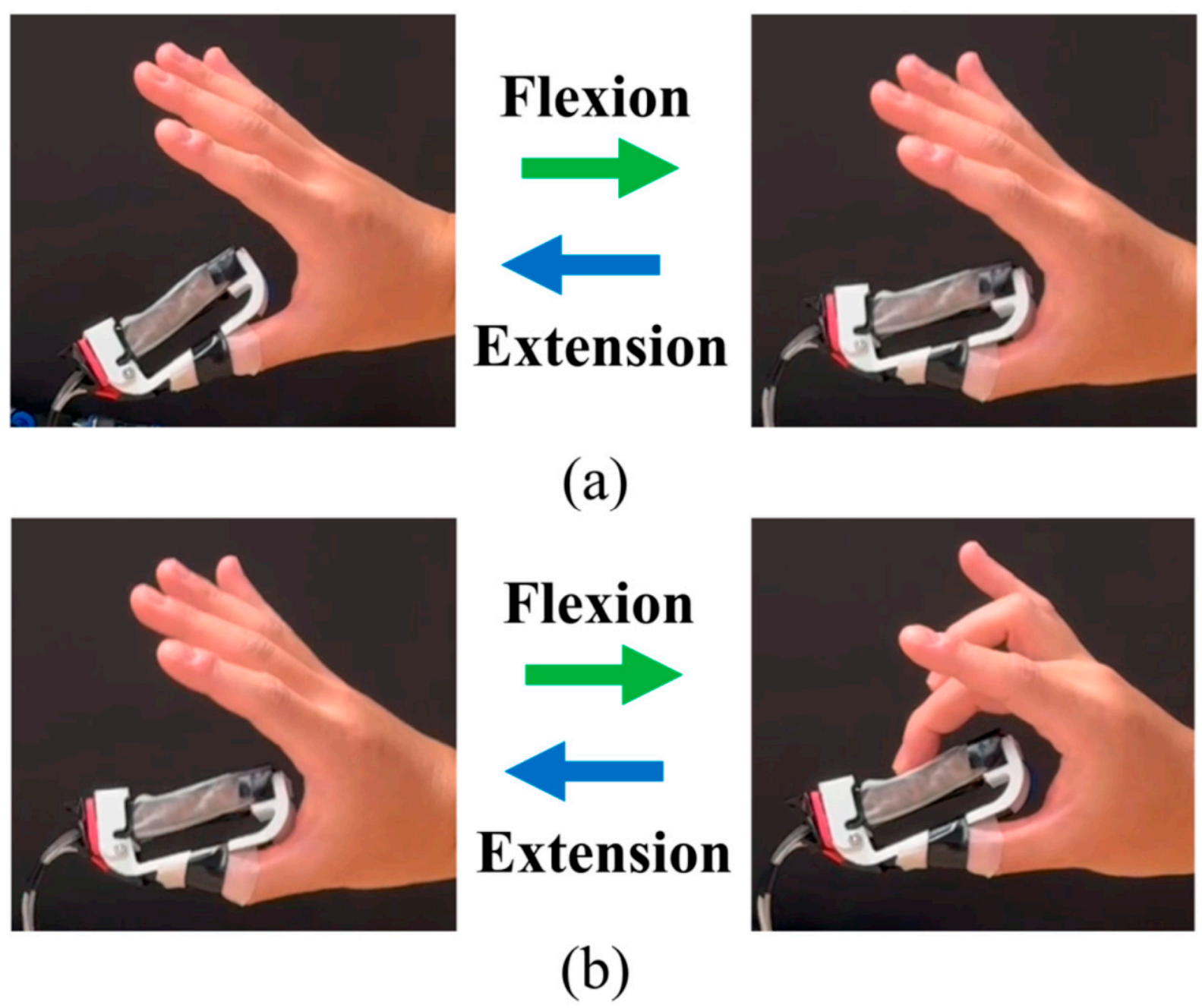

**Figure 10.** Demonstration of finger mobility. (**a**) Flexion/extension of the interphalangeal joint of the thumb; (**b**) flexion/extension of the middle finger.

Alongside device development, we are preparing to conduct user studies with multiple healthy participants to validate the functionality and ergonomics of the HJ-Haptic. While one of the authors participated in an initial teleoperation experiment, formal user studies were not conducted at this stage.

These studies will determine whether users can perceive different levels of object stiffness with the HJ-Haptic. Nevertheless, we expect that users can differentiate the object stiffness, given that the HJ-Haptic has a stiffness-changing ratio per kPa of 0.077, higher than other jamming haptic devices in Table 3. Additionally, human operators perceive object stiffness through force feedback, a combination of stiffness and indentation. The HJ-Haptic renders stiffness and provides force feedback with finger indentation. For instance, the HJ-Haptic provides force feedback values of 11.5 N and 26.4 N with a 10 mm indentation at vacuum pressures of 0 kPa and 30 kPa, respectively. We believe that a human fingertip can easily detect this significant force difference (14.9 N). However, formal user studies are required to confirm this expectation and evaluate the user experience.

Additionally, the user studies should assess user comfort and usability over extended periods. Participants will wear the device for prolonged durations while performing object-grasping tasks, and then a detailed questionnaire and motion analysis will be conducted to evaluate perceived comfort, usability, and any long-term effects on finger mobility.

As for potential applications, the HJ-Haptic can be applied beyond teleoperation to extended reality (XR), such as virtual reality (VR) [28] and augmented reality (AR) [30]. XR applications face similar technical challenges, including the rendering stiffness of virtual objects [38] and the presence of network communication latency [39,40], which can impact the quality of haptic sensation and the overall virtual experience. The stiffness-rendering capability of the HJ-Haptic presents a promising solution to these challenges in XR settings. In the future, we plan to explore the potential of the HJ-Haptic to deliver immersive haptic feedback for XR applications. Additionally, we will conduct user studies to evaluate its effectiveness in rendering object stiffness and force sensations during virtual object-grasping tasks, aiming to enhance the user experience.

## 5. Conclusions

This paper presents a wearable, lightweight (20 g), and compact haptic device, the HJ-Haptic, designed for bilateral teleoperation. Its stiffness can be varied up to 2.3 times with 30 kPa, with stiffness controlled linearly by vacuum pressure. This device can be employed in robot teleoperation to render object stiffness in object-grasping tasks.

Integrating the HJ-Haptic into a bilateral teleoperation framework allows for real-time kinesthetic feedback. We validated this framework through teleoperated object-grasping experiments with a small RMSE and strong correlations in terms of teleoperated motion, stiffness rendering, and interaction force feedback. This allows operators to feel remote objects' stiffness and gauge their grip force.

Future formal user studies should be performed to evaluate the user experience. Ultimately, with a further development, the HJ-Haptic may enable operators to control the remote gripper intuitively and safely, with a reliable haptic feedback of remote object stiffness.

**Author Contributions:** Conceptualization, T.M.K. and W.T.C.; data curation, T.M.K.; formal analysis, T.M.K.; funding acquisition, W.T.C.; investigation, T.M.K., B.Z. and W.T.C.; methodology, T.M.K. and W.T.C.; project administration, B.Z.; software, T.M.K.; supervision, W.T.C.; visualization, T.M.K.; writing—original draft, T.M.K.; and writing—review and editing, T.M.K., B.Z., and W.T.C. All authors have read and agreed to the published version of the manuscript.

**Funding:** This research was funded by the Agency for Science, Technology, and Research (A*STAR) under its IAF-ICP Programme I2001E0067 and the Schaeffler Hub for Advanced Research at NTU.

**Institutional Review Board Statement:** Not applicable.

**Informed Consent Statement:** Not applicable.

**Data Availability Statement:** The original contributions presented in this study are included in the article Further inquiries can be directed to the corresponding author.

**Acknowledgments:** The authors acknowledge the use of ChatGPT (OpenAI, Version GPT-4) for grammar and language checks in the preparation of this manuscript.

**Conflicts of Interest:** The authors declare no conflicts of interest.